\documentclass[conference]{IEEEtran}
\IEEEoverridecommandlockouts

\usepackage{cite}
\usepackage{amsmath,amssymb,amsfonts}
\usepackage{algorithmic}
\usepackage{graphicx}
\usepackage{textcomp}
\usepackage{xcolor}
\usepackage{url}
\usepackage{booktabs}
\usepackage{array}
\usepackage{soul}
\usepackage{tabularx}
\usepackage{adjustbox}
\usepackage[table]{xcolor}
\usepackage[hidelinks, colorlinks=true, linkcolor=blue, citecolor=blue, breaklinks=true]{hyperref}
\def\BibTeX{{\rm B\kern-.05em{\sc i\kern-.025em b}\kern-.08em
    T\kern-.1667em\lower.7ex\hbox{E}\kern-.125emX}}
\begin{document}

\title{Semantic Class Distribution Learning for Debiasing Semi-Supervised Medical Image Segmentation
}



















\author{
\IEEEauthorblockN{
Yingxue Su\textsuperscript{2}, 
Yiheng Zhong\textsuperscript{1}\dag, 
Keying Zhu\textsuperscript{3}, 
Zimu Zhang\textsuperscript{3}, 
Zhuoru Zhang\textsuperscript{3},\\ 
Yifang Wang\textsuperscript{4},
YuXin Zhang\textsuperscript{5}, 
Xinyuan Zheng\textsuperscript{1}, 
Jingxin Liu\textsuperscript{3}\dag, 
Xiaofeng Liu\textsuperscript{1}\dag
}

\IEEEauthorblockA{
\IEEEauthorblockA{\textsuperscript{1} Yale University, United States}
\IEEEauthorblockA{\textsuperscript{2} Carnegie Mellon University, United States}
\IEEEauthorblockA{\textsuperscript{3} Xi'an Jiaotong-Liverpool University, China}
\IEEEauthorblockA{\textsuperscript{4} University College London, London, United Kingdom}
\IEEEauthorblockA{\textsuperscript{5} Wuhan University, China}
\IEEEauthorblockA{Emails: yiheng.zhong@yale.edu, jingxin.liu@xjtlu.edu.cn, xiaofeng.liu@yale.edu}
}

\thanks{\textsuperscript{\dag}Corresponding Author.}

}


\maketitle

\begin{abstract}
Medical image segmentation is critical for computer-aided diagnosis. However, dense pixel-level annotation is time-consuming and costly, and medical datasets often exhibit severe class imbalance. Such an imbalance causes minority structures to be overwhelmed by dominant classes in feature representations, hindering the learning of discriminative features and making reliable segmentation particularly challenging. To address this, we propose the Semantic Class Distribution Learning (SCDL) framework, a plug-and-play module that mitigates supervision and representation biases by learning structured class-conditional feature distributions. SCDL integrates Class Distribution Bidirectional Alignment (CDBA) to align embeddings with learnable class proxies and leverages Semantic Anchor Constraints (SAC) to guide proxies using labeled data. Experiments on the Synapse and AMOS datasets demonstrate that SCDL largely improves segmentation performance across both overall and class-level metrics, with particularly notable gains for several low-frequency organs. Our anonymous code is released at \url{https://anonymous.4open.science/r/SCDL}.
\end{abstract}


\section{Introduction}
\label{sec:intro}
Accurate and reliable medical image segmentation is crucial for computer-aided diagnosis, treatment, and prognosis. However, dense pixel-level annotation is time-consuming and costly\cite{yu2019uncertainty,chen2024transunet,qi2023mdf,tang2019clinically}. Therefore, Semi-Supervised Medical Image Segmentation (SSMIS) has attracted widespread attention, as it typically leverages unlabeled data to alleviate annotation burden through strategies such as consistent regularization, pseudo-labeling and contrastive representation learning\cite{chen2023magicnet,yu2019uncertainty,rizve2021defense,XU2023102880,LUO2022102517}. However, in real-world scenarios, medical segmentation often suffers from class imbalance.
This imbalance causes the training process to be dominated by large structures. Under conditions of scarce annotations, small structures are difficult to segment in a stable and reliable manner\cite{qiao2022semi,yu2019uncertainty,11356599,xiang2022fussnet}.

\begin{figure}[t]
  \centering
  \includegraphics[width=\linewidth]{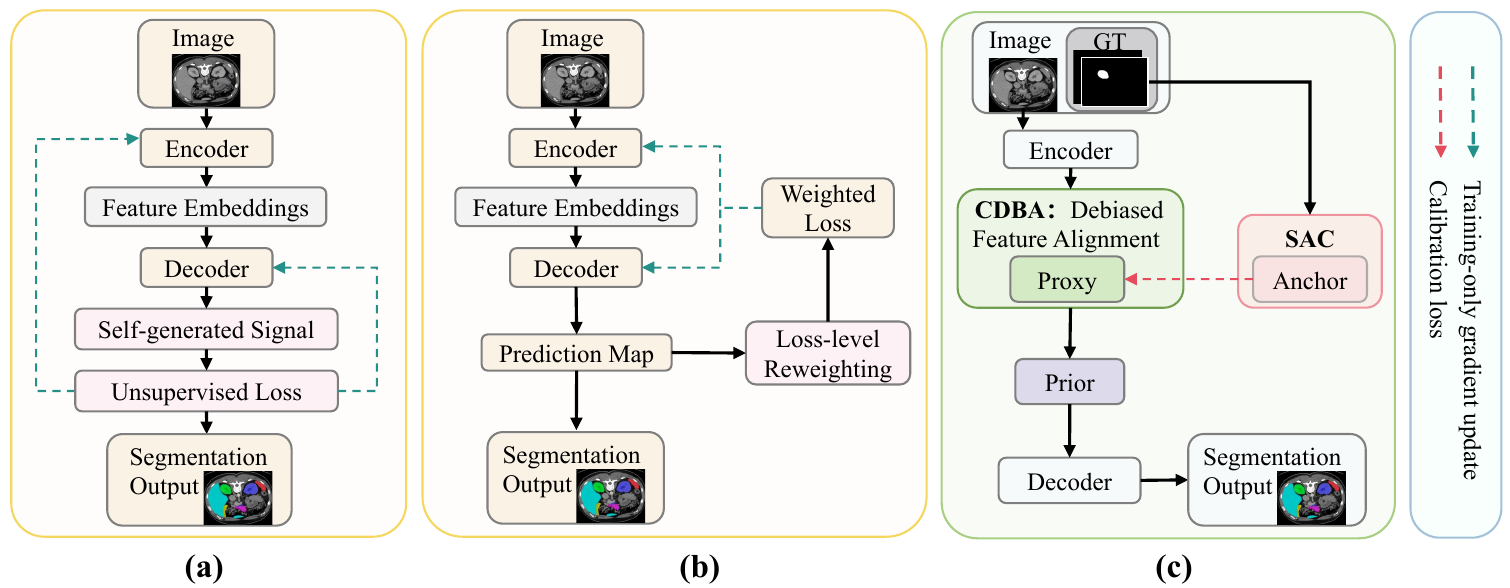}
  \caption{Comparison of imbalance-aware SSMIS paradigms: (a) Self-generated supervision uses model-derived unlabeled signals and may amplify head-class bias. (b) Loss reweighting and output calibration rebalance training but do not constrain class-conditional features. (c) SCDL debiases feature distributions via CDBA, distribution priors, and SAC anchoring.
}
  \label{picture1_label}
\end{figure}

Under such a pixel-level long-tailed distribution, the combination of class imbalance and semi-supervised mechanisms gives rise to two major challenges. As shown in Fig.~\ref{picture1_label}, the first challenge is that the supervision signal becomes biased toward head classes. Since large organs occupy more pixels, the model’s gradient updates are more likely to favor these classes. Many SSMIS methods rely on self-generated signals from unlabeled data, such as soft pseudo labeling and consistency constraints, which further reinforce the learning of head classes and result in insufficient training for tail classes \cite{qiao2022semi,yu2019uncertainty,lu2023uncertainty}. Second, this bias leads to an imbalance at the representation level. Existing methods often mitigate imbalance through techniques such as reweighting or output alignment, but these approaches mainly operate at the loss or output level and lack direct constraints on class-conditional feature distributions \cite{lin2022calibrating}. As a result, features of head classes become more compact, while features of tail classes drift toward regions dominated by head classes, which blurs class boundaries and degrades segmentation performance on small structures \cite{tang2025density,11356599,taghanaki2019combo}. Notably, many existing methods use unlabeled data mainly for local consistency regularization and rarely exploit it to explicitly correct the skewness of class-conditional feature distributions
\cite{su2024mutual,yao2022enhancing,li2020self}. Therefore, unlabeled data may be underutilized for shaping minority-class feature distributions, allowing imbalance-induced representation bias to persist.

\begin{figure*}[!t]
\centering
\includegraphics[width=0.9\textwidth]{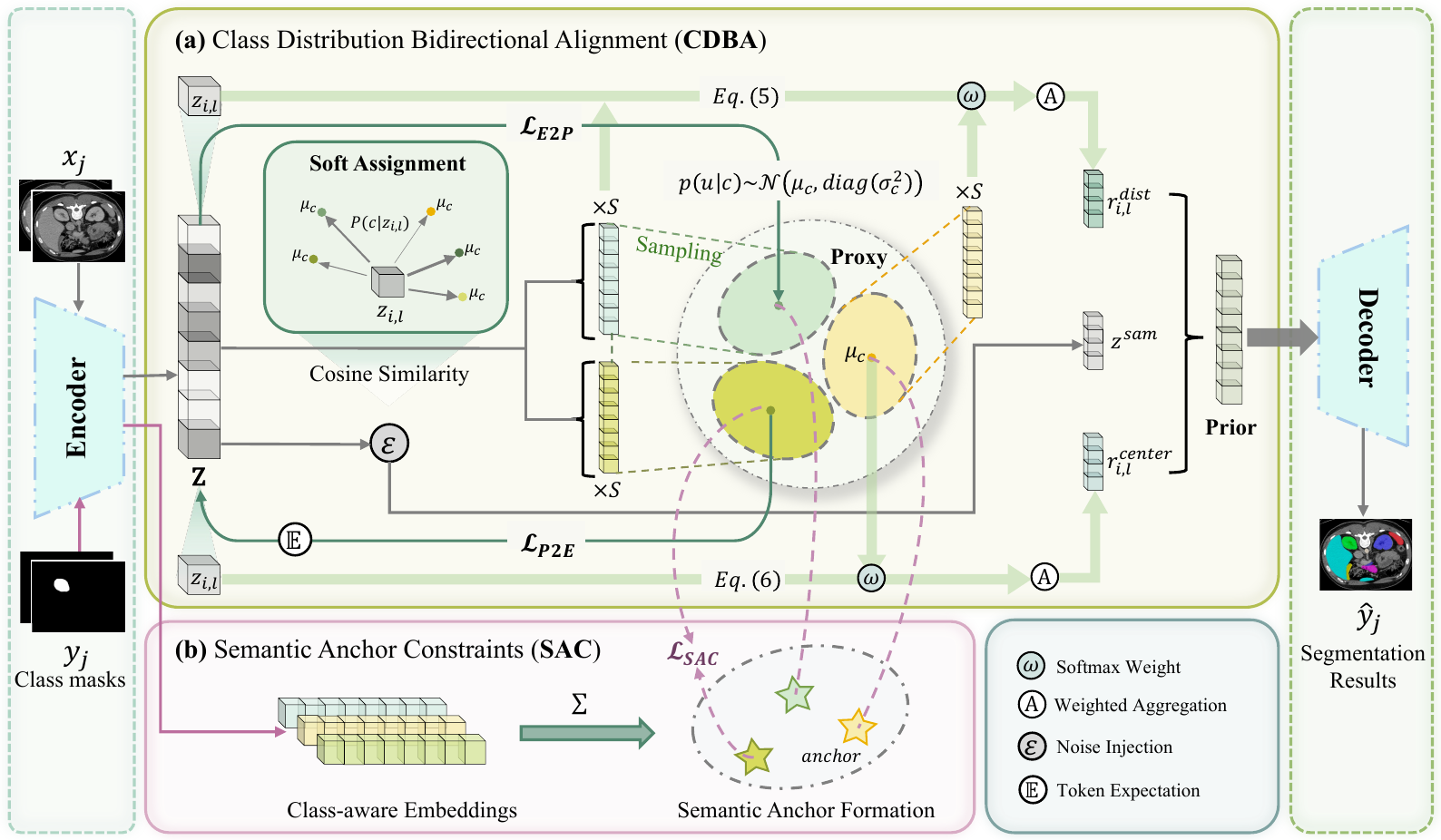}
\caption{\textbf{Overview of SCDL.} 
(a) \textbf{CDBA} learns class proxy distributions and aligns them with token features to generate proxy-guided priors for downstream tasks. 
(b) \textbf{SAC} extracts semantic anchors from labeled regions and uses them to regularize the proxies, ensuring consistent class semantics across categories.}
\label{picture2_label}
\end{figure*}

To address the above limitations, we propose a Semantic Class Distribution Learning (SCDL) framework, as illustrated in Fig.~\ref{picture2_label}. The framework is integrated into existing segmentation networks as a plug-and-play module. While fully leveraging labeled data to provide semantic supervision, it guides unlabeled data to participate in learning at the distribution level, thereby learning structured class-conditional feature distributions in the embedding space. This explicitly reshapes the class-conditional feature structure and mitigates both supervision bias and representation-level imbalance. Class Distribution Bidirectional Alignment (CDBA) models each semantic class as a learnable proxy distribution and enforces bidirectional alignment constraints between embeddings and the proxy distributions. This mechanism is not affected by differences in sample scale across classes and enables stable modeling of minority class distributions while providing consistent learning signals. Semantic Anchor Constraints (SAC) align each proxy distribution with semantic anchors constructed from labeled data, providing reliable semantic supervision to ensure that the proxy distributions capture the true semantics of each class. This process refines the supervision signal, corrects learned representations, and prevents bias from class frequency imbalance, thereby ensuring consistent feature representations.

Our contributions are summarized as follows: 
(i) We propose the SCDL framework, which mitigates coupled supervision bias and representation-level imbalance by learning structured class-conditional distributions. (ii) We introduce CDBA, which shapes class-conditional structures through bidirectional alignment between embeddings and proxy distributions. (iii) We propose SAC, which uses labeled data to construct semantic anchors, guiding proxy distributions toward true class semantics. (iv) Experiments on the Synapse and AMOS datasets show that SCDL achieves superior results in overall and tail-class segmentation.

\section{Related Work}

\subsection{Semi-supervised Medical Image Segmentation under Class Imbalance}

Semi-supervised medical image segmentation reduces the reliance on voxel-level annotations by exploiting unlabeled images. A common strategy is to convert unlabeled data into auxiliary supervision through consistency regularization, teacher-student learning, pseudo-labeling, or co-training \cite{yu2019uncertainty,rizve2021defense,su2024mutual,lee2013pseudo,ijcai2018p278,ijcai2023p467}. For example, UA-MT extends the mean-teacher paradigm to semi-supervised 3D medical segmentation by using uncertainty estimation to suppress unreliable consistency targets \cite{yu2019uncertainty}. MagicNet further introduces a magic-cube partition-and-recovery strategy for semi-supervised multi-organ segmentation, where anatomical location and size priors are used to guide consistency learning across labeled and unlabeled volumes \cite{chen2023magicnet}. Together, these methods show that unlabeled images can provide effective supervision. However, when anatomical structures are highly imbalanced in size, model-generated supervision may become uneven across classes.

This issue has motivated a series of studies on class-imbalanced medical segmentation under limited supervision. Existing methods address the problem from several perspectives, including supervision calibration, debiased training, optimization adjustment, decoder rebalancing, and semantic-prior incorporation. CLD exploits label quantity and position distributions to calibrate training in class-imbalanced barely-supervised segmentation \cite{lin2022calibrating}. DHC introduces dual-debiased heterogeneous co-training to reduce both data and learning biases \cite{wang2023dhc}, whereas GA analyzes class-dependent gradient bias and adjusts optimization for tail classes \cite{qi2024gradient}. More recently, SKCDF improves small-organ learning through decoupled data flows, semantic knowledge complementarity, and an auxiliary balanced segmentation head \cite{zhang2025semantic}. Collectively, these studies indicate that class imbalance in SSMIS is closely associated with supervision generation and optimization dynamics, rather than being merely a loss-design issue. Nevertheless, their correction mechanisms are mainly implemented through label-distribution calibration, co-training debiasing, gradient adjustment, decoder rebalancing, or semantic-prior incorporation, thereby primarily intervening in the supervision source, optimization process, prediction head, or auxiliary semantic guidance. However, the representation-level organization of class-conditional features, particularly for tail classes, remains insufficiently explored.

\subsection{Representation Learning for Class-conditional Distribution Alignment}

This limitation motivates feature organization beyond loss- or output-level rebalancing, as such methods adjust losses, predictions, or pseudo-label distributions but provide limited control over the geometry of class-specific features. To improve feature separability with limited annotations, contrastive representation learning has been introduced into semi-supervised medical segmentation. SimCVD learns structural representations through voxel-wise contrastive representation distillation \cite{you2022simcvd}, while density-aware contrastive learning exploits neighborhood density in the feature space to improve intra-class compactness \cite{tang2025density}. For class-imbalanced 3D medical segmentation, SuVCL further enhances local feature discriminability through sub-volume contrastive learning \cite{xu2025suvcl}. Although these methods exploit feature-space supervision beyond final logits, they mainly regulate voxel-wise, sample-wise, or region-wise relationships and lack explicit class-level references for distribution-aware alignment under class imbalance.

Different from such relation-based feature learning, prototype- and proxy-based methods organize the embedding space with class-level references, such as prototype centers or learnable proxies \cite{snell2017prototypical,movshovitz2017proxy,kim2020proxyanchor}. In semi-supervised medical segmentation, prototypes have also been used for consistency learning, feature compactness, and uncertainty-aware voxel-prototype matching \cite{zhang2023selfaware,yuan2025effective}. Despite providing semantic guidance beyond individual pixel- or voxel-level features, most prototype- or proxy-based methods still represent each class with compact centers or prototype-matching targets, which may be insufficient to characterize intra-class variations and uncertainty under class imbalance. This limitation becomes more pronounced for small organs, where reliable samples are limited and unlabeled tail-class features are often sparse or unreliable, making estimated prototype centers susceptible to bias from dominant organs or noisy pseudo-labeled voxels. Moreover, matching features to estimated or learned references imposes point-wise or target-matching constraints, rather than explicitly enforcing distribution-level alignment between class-level references and feature embeddings from both labeled and unlabeled data. This motivates SCDL, which learns class-level proxy distributions anchored by labeled semantic cues and performs bidirectional distribution-level alignment between proxies and feature embeddings.

\section{Methodology}

\subsection{Class Distribution Bidirectional Alignment (CDBA)}
\label{sec:cdba}
To mitigate supervision and representation bias toward head classes, CDBA learns structured class-conditional distributions in the embedding space and enforces bidirectional alignment between embeddings and distributions, enabling minority classes to receive consistent supervision even under sparse annotations. 

\noindent\textbf{Class Distribution Modeling.} Each semantic class $c \in \{1, \dots, C\}$ is represented by a learnable proxy distribution in the embedding space:
\begin{equation}
p(u|c) = \mathcal{N}(\mu_c, \mathrm{diag}(\sigma_c^2)),
\end{equation}
where $\mu_c$ denotes the mean vector of class $c$, and $\sigma_c$ denotes the standard deviation vector for each dimension; the corresponding diagonal covariance matrix is $\mathrm{diag}(\sigma_c^2)$; both $\mu_c$ and $\sigma_c$ are randomly initialized and parameterized as trainable proxies. Given a batch of embeddings $\mathbf{Z} \in \mathbb{R}^{B \times L \times D}$, we compute a soft assignment of each token embedding to all class proxies:
\begin{equation}
P(c|z_{i,l}) = \mathrm{softmax}_c\big(\cos(z_{i,l}, \mu_c)\big),
\end{equation}
where $z_{i,l}\in\mathbb{R}^{D}$ denotes the $l$-th token embedding of the $i$-th sample in the mini-batch, and $\cos(\cdot,\cdot)$ is the cosine similarity. This soft assignment enables each embedding to associate with multiple classes, alleviating supervision bias and allowing gradients to flow to minority-class proxies even under sparse labels. To further prevent minority-class embeddings from being dominated by majority distributions, we introduce bidirectional distribution alignment to explicitly align embeddings with their corresponding class distributions:

\noindent\textbf{(i)Embedding-to-Proxy (E2P) Alignment.} Each embedding is encouraged to move closer to its soft-assigned proxy distributions:
\begin{equation}
\mathcal{L}_{E2P} = \sum_{i=1}^{B} \sum_{l=1}^{L} \sum_{c=1}^{C} P(c|z_{i,l}) \cdot \big[ 1 - \cos(z_{i,l}, \mu_c) \big].
\end{equation}
By weighting cosine distance with soft assignment probabilities, $\mathcal{L}_{E2P}$ encourages embeddings to align with class proxies. This soft weighting allows each embedding to influence multiple proxies according to its assignment weights, promoting balanced gradient flow and mitigating majority-class dominance.


\noindent\textbf{(ii) Proxy-to-Embedding (P2E) Alignment.}
While E2P encourages embeddings to move toward their softly assigned proxies, P2E further optimizes each proxy to discriminate embeddings that are semantically related to it from those that are not. Specifically, for each class proxy $\mu_c$, we define a soft positive similarity and a soft negative similarity as:
\begin{equation}
\begin{aligned}
\mathcal{E}_c^{+}
&=
\frac{
\sum_{i=1}^{B}\sum_{l=1}^{L}
P(c|z_{i,l}) \, \cos(z_{i,l}, \mu_c)
}{
\sum_{i=1}^{B}\sum_{l=1}^{L} P(c|z_{i,l}) + \delta
},\\[1mm]
\mathcal{E}_c^{-}
&=
\frac{
\sum_{i=1}^{B}\sum_{l=1}^{L}
\big(1-P(c|z_{i,l})\big) \, \cos(z_{i,l}, \mu_c)
}{
\sum_{i=1}^{B}\sum_{l=1}^{L} \big(1-P(c|z_{i,l})\big) + \delta
},
\end{aligned}
\end{equation}
where $\delta$ is a small constant for numerical stability. The P2E loss is then formulated as:
\begin{equation}
\mathcal{L}_{\mathrm{P2E}}
=
\frac{1}{C}
\sum_{c=1}^{C}
\exp\Big(
-\big(\mathcal{E}_c^{+}-\mathcal{E}_c^{-}\big)
\Big).
\end{equation}
Here, $\mathcal{E}_c^{+}$ measures the average similarity between proxy $\mu_c$ and embeddings softly assigned to class $c$, while $\mathcal{E}_c^{-}$ measures its average similarity to embeddings that are unlikely to belong to class $c$. By minimizing $\mathcal{L}_{\mathrm{P2E}}$, each proxy is encouraged to increase its similarity with class-relevant embeddings and decrease its similarity with class-irrelevant embeddings, thereby improving proxy discriminability and reducing the dominance of majority-class embeddings.

\noindent\textbf{Proxy Sampling and Feature Enrichment.}
To exploit the learned class-conditional proxy distributions, we construct token-wise priors via lightweight proxy sampling, providing structured semantic guidance under distributional uncertainty.

\noindent\textbf{(i) Distribution-Weighted Prior.} 
For each embedding $z_{i,l}$, we sample $S$ variance-aware proxies from each class distribution using the reparameterization trick:
\begin{equation}
u_c^s = \mu_c + \sigma_c \odot \epsilon_c^s,
\quad 
\epsilon_c^s \sim \mathcal{N}(0, I),
\quad 
s=1,\dots,S,
\end{equation}
where $\epsilon_c^s$ is a standard Gaussian noise vector and $\odot$ denotes element-wise multiplication. These samples are used to compute distribution-level token-to-class similarities:
\begin{equation}
\mathbf{r}^{dist}_{i,l} = \sum_{c=1}^{C} 
\frac{\exp\Big(\frac{1}{S}\sum_{s=1}^S \cos(z_{i,l}, u_c^s)\Big)}
{\sum_{c'=1}^{C} \exp\Big(\frac{1}{S}\sum_{s=1}^S \cos(z_{i,l}, u_{c'}^s)\Big)}
\, \mu_c
\in \mathbb{R}^D.
\end{equation}
Although $\mathbf{r}^{dist}_{i,l}$ is represented as a weighted combination of proxy means, its weights are estimated from variance-aware samples. Thus, $\sigma_c$ controls the sampling range of each class proxy and is optimized through the differentiable similarity estimation, enabling the prior to capture distribution-level uncertainty while preserving soft multi-class associations.

\noindent\textbf{(ii) Center-Similarity Prior.} 
In contrast, the center-similarity prior aligns each token directly to the class means:
\begin{equation}
\mathbf{r}^{center}_{i,l} = \sum_{c=1}^{C} 
\frac{\exp\big(\cos(z_{i,l}, \mu_c)\big)}
{\sum_{c'=1}^{C} \exp\big(\cos(z_{i,l}, \mu_{c'})\big)}
\, \mu_c
\in \mathbb{R}^D.
\end{equation}
Unlike the distribution-weighted prior, this mean-based reweighting ignores variance but provides complementary token-to-class signals, improving representation stability and discriminability.
Additionally, we perform local perturbation sampling around each token with a learnable noise scale, aggregating normalized samples to obtain a token sampling prior $\mathbf{z}^{\mathrm{sam}}\in\mathbb{R}^{B\times L\times D}$, which enhances robustness.
Finally, the three priors are concatenated along the feature dimension:
\begin{equation}
\mathbf{z}^{\mathrm{prior}}=\mathrm{concat}\big(\mathbf{r}^{dist},\ \mathbf{r}^{center},\ \mathbf{z}^{\mathrm{sam}}\big).
\end{equation}
Then, a lightweight projection matches the decoder channel size, and the prior is resized for additive injection at each decoder stage, allowing the learned distributional cues to guide segmentation.

\subsection{Semantic Anchor Constraints (SAC)}
\label{sec:sac}
Although CDBA learns class-conditional distributions and enforces bidirectional alignment between embeddings and proxies, the proxies are randomly initialized and lack direct semantic supervision, leaving their correspondence to true class semantics unconstrained. 
To address this issue, we introduce Semantic Anchor Constraints (SAC), which provide high-level semantic guidance for proxy learning. SAC assumes that encoder embeddings from labeled regions contain sufficient class-specific semantic information. These embeddings are therefore used to guide the learning of corresponding class proxies, promoting intra-class alignment and enhancing inter-class separability.

\definecolor{HeaderGreen}{HTML}{87C1AA}
\definecolor{VNetRow}{HTML}{A8B8A0}
\begin{table}[!thbp]
\centering
\caption{Comparison of quantitative results with state-of-the-art models on Synapse (20\% labeled) and AMOS (5\% labeled).}
\label{tab:dsc_asd_summary}

\begingroup
\scriptsize
\renewcommand{\arraystretch}{0.90}
\setlength{\tabcolsep}{1.6pt}

\begin{adjustbox}{max width=\columnwidth}
\begin{tabularx}{\columnwidth}{
  >{\raggedright\arraybackslash}p{1.85cm}
  >{\centering\arraybackslash}X
  >{\centering\arraybackslash}X
  >{\centering\arraybackslash}X
  >{\centering\arraybackslash}X
}
\toprule

\rowcolor{HeaderGreen}
\textbf{Methods} &
\multicolumn{2}{c}{\textbf{Synapse}} &
\multicolumn{2}{c}{\textbf{AMOS}} \\
\cmidrule(lr){2-3}\cmidrule(lr){4-5}
& DSC$\uparrow$ & ASD$\downarrow$ & DSC$\uparrow$ & ASD$\downarrow$ \\
\midrule

\rowcolor{VNetRow}
VNet (fully)~\cite{milletari2016v} 
& 68.49$\pm$3.5 & 6.08$\pm$5.4 
& 76.50$\pm$2.1 & 2.01$\pm$3.2 \\
\midrule

CLD~\cite{lin2022calibrating} 
& 49.47$\pm$2.9 & 34.73$\pm$7.6 
& 46.10$\pm$2.5 & 15.86$\pm$4.3 \\

DHC~\cite{wang2023dhc} 
& 47.88$\pm$1.9 & 8.73$\pm$2.7 
& 48.69$\pm$0.3 & 9.84$\pm$0.5 \\

SimiS \cite{chen2022embarrassingly} 
& 50.45$\pm$2.7 & 33.11$\pm$3.6 &47.27$\pm$1.6 & 11.51$\pm$2.3 \\

SKCDF~\cite{zhang2025semantic} 
& 55.10$\pm$1.7 & 6.28$\pm$1.4
& 52.56$\pm$0.7 & 6.21$\pm$0.2 \\

GenericSSL~\cite{wang2023towards} 
& 60.37$\pm$0.5 & 7.13$\pm$1.7
& 52.63$\pm$3.5 & 8.74$\pm$2.9\\

GA-MagicNet~\cite{qi2024gradient} 
& 66.00$\pm$1.9 & 3.42$\pm$0.6 
& 59.15$\pm$1.5 & 8.66$\pm$1.3 \\

GA-CPS~\cite{qi2024gradient} 
& 66.29$\pm$0.3 & 5.44$\pm$0.4 
& 50.90$\pm$0.5 & 13.77$\pm$0.6 \\

\midrule

\textbf{DHC+SCDL} 
& 50.67$\pm$1.7 & 4.58$\pm$0.8 
& 50.44$\pm$0.3 & 9.86$\pm$0.3 \\

\textbf{GA-MagicNet+SCDL} 
& 66.75$\pm$1.3 & 3.65$\pm$0.3 
& \textbf{62.16$\pm$1.2} & \textbf{5.65$\pm$0.4} \\

\textbf{GA-CPS+SCDL} 
& \textbf{67.50$\pm$0.9} & \textbf{3.32$\pm$0.8} 
& 61.57$\pm$1.0 & 10.08$\pm$1.1 \\

\bottomrule
\end{tabularx}
\end{adjustbox}
\endgroup
\end{table}

\noindent\textbf{Semantic Anchor Formation.} For each class, we identify class-specific regions based on ground-truth masks and mask out all non-target pixels. The masked class-specific images are passed through the encoder shared with the original image to obtain class-aware embeddings. 


Formally, let $\mathcal{Z}_c=
\left\{
\hat{z}_{i,l}^{(c)}
\mid
(x_i,y_i)\in\mathcal{D}_{\mathrm{L}},
\ \tilde{m}_{i,l}^{(c)}=1
\right\}$
denote the set of class-masked token embeddings for class $c$, where $\tilde{m}{i,l}^{(c)}$ indicates whether the token corresponding to $\hat{z}{i,l}^{(c)}$ belongs to class $c$. The semantic anchor is then computed as:
\begin{equation}
\text{anchor}_c = \frac{1}{|\mathcal{Z}_c|} \sum_{z \in \mathcal{Z}_c}\hat{z}.
\end{equation}

During backpropagation, semantic anchors are detached to prevent gradients from updating the encoder, ensuring that the SAC loss only updates the proxies.

\begin{table*}[!t]
\centering
\caption{Per-class Dice scores (\%) on the Synapse dataset with 20\% labeled data.}
\label{tab:synapse_20p}

\begingroup
\fontsize{8}{9.2}\selectfont
\setlength{\tabcolsep}{0.75pt}
\renewcommand{\arraystretch}{0.85}

\begin{tabular*}{\textwidth}{@{\extracolsep{\fill}}@{}p{0pt}p{3.3cm}*{13}{c}@{}}
\toprule
\rowcolor{HeaderGreen}
\multicolumn{2}{c}{Methods} &
\multicolumn{13}{c}{Average Dice of Each Class} \\
\cmidrule(lr){1-2}\cmidrule(lr){3-15}
& &
Sp & RK & LK & Ga & Es & Li & St & Ao & IVC & PSV & PA & RAG & LAG \\
\midrule

\rowcolor{VNetRow}
\multicolumn{2}{l}{VNet (fully)~\cite{milletari2016v}} &
90.2 & 91.9 & 90.7 & 38.3 & 30.9 & 94.8 & 75.6 & 79.1 & 81.4 & 62.1 & 48.5 & 48.9 & 58.0 \\
\midrule

& CLD~\cite{lin2022calibrating} &
83.3 & 86.7 & 85.7 & 1.3 & 0.0 & 85.9 & 49.1 & 74.5 & 76.3 & 52.4 & 33.8 & 14.1 & 0.0 \\

& DHC~\cite{wang2023dhc} &
    72.5 & 55.2 & 49.9 & \textbf{60.3} & \textbf{50.0} & 87.8 & 31.5 & 67.0 & 54.5 & 38.7 & 27.8 & 19.4 & 7.8 \\

& SimiS \cite{chen2022embarrassingly} &
83.3 & 90.8 & 85.8 & 9.2 & 0.0 & 85.6 & 55.0 & 73.6 & 71.7 & 50.4 & 34.0 & 0.0 & 16.6 \\

& SKCDF~\cite{zhang2025semantic} &
85.5 & 85.6 & 82.4 & 17.3 & 32.8 & 87.6 & 45.3 & 70.7 & 70.4 & 44.9 & 32.5 & 30.3 & 31.1 \\

& GenericSSL~\cite{wang2023towards} &
85.2 & 85.8 & 84.1 & 19.0 & 37.3 & 89.9 & 51.0 & 76.9 & 74.2 & 48.9 & 46.4 & 39.7 & 46.2 \\

& GA-MagicNet~\cite{qi2024gradient} &
79.4 & \textbf{92.2} & 89.5 & 24.0 & 48.1 & 90.7 & 52.9 & \textbf{80.7} & 81.2 & 64.6 & 43.3 & 48.5 & \textbf{62.9} \\

& GA-CPS~\cite{qi2024gradient} &
85.5 & 90.1 & 88.3 & 26.7 & 40.2 & 92.7 & \textbf{64.8} & 79.7 & 79.1 & \textbf{66.8} & 45.5 & 44.7 & 57.4 \\

\midrule


& \textbf{DHC+SCDL} &
79.1 & 55.6 & 48.8 & 52.8 & 48.9 & 88.8 & 45.0 & 75.6 & 59.3 & 38.1 & 24.0 & 32.1 & 10.8 \\

& \textbf{GA-MagicNet+SCDL} &
79.3 & 91.5 & \textbf{90.3} & 20.3 & 49.8 & 88.9 & 63.5 & 79.0 & 80.8 & 64.6 & 46.5 & \textbf{51.6} & 61.6 \\

& \textbf{GA-CPS+SCDL} &
\textbf{88.2} & 91.7 & 89.3 & 25.4 & 38.7 & \textbf{93.7} & 64.4 & 78.0 & \textbf{81.8} & 65.4 & \textbf{49.4} & 49.2 & 62.4 \\

\bottomrule
\end{tabular*}

\endgroup
\end{table*}

\definecolor{HeaderGreen}{HTML}{87C1AA}
\definecolor{VNetRow}{HTML}{A8B8A0}
\begin{table*}[!t]
\centering
\caption{Per-class Dice scores (\%) on the AMOS dataset with 5\% labeled data.}
\label{tab:amos_5p}

\begingroup
\fontsize{8}{9.2}\selectfont
\setlength{\tabcolsep}{0.75pt}
\renewcommand{\arraystretch}{0.85}

\begin{tabular*}{\textwidth}{@{\extracolsep{\fill}}@{}p{0pt}
>{\raggedright\arraybackslash}p{3.2cm}
*{15}{>{\centering\arraybackslash}m{0.57cm}@{\hskip 0.7pt}}
@{}}
\toprule
\rowcolor{HeaderGreen}
\multicolumn{2}{c}{Methods} &
\multicolumn{15}{c}{Average Dice of Each Class} \\
\cmidrule(lr){1-2}\cmidrule(lr){3-17}
& &
Sp & RK & LK & Ga & Es & Li & St & Ao & IVC & PA & RAG & LAG & Du & Bl & P/U \\
\midrule

\rowcolor{VNetRow}
\multicolumn{2}{l}{VNet (fully)~\cite{milletari2016v}} &
92.2 & 92.2 & 93.3 & 65.5 & 70.3 & 95.3 & 82.4 & 91.4 & 85.0 & 74.9 & 58.6 & 58.1 & 65.6 & 64.4 & 58.3 \\
\midrule

& CLD~\cite{lin2022calibrating} &
67.2 & 68.5 & 71.4 & 41.0 & 21.0 & 76.1 & 42.4 & 69.8 & 52.1 & 37.9 & 24.7 & 23.4 & 22.7 & 38.1 & 35.2 \\


& DHC~\cite{wang2023dhc} &
73.3 & 63.8 & 62.6 & 40.3 & 34.4 & 81.8 & 47.2 & 69.0 & 56.9 & 45.5 & 30.5 & 16.8 & 31.8 & 49.2 & 27.7 \\

& SimiS \cite{chen2022embarrassingly} &
77.4 & 72.5 & 68.7 & 32.1 & 14.7 & 86.6 & 46.3 & 74.6 & 54.2 & 41.6 & 24.4 & 17.9 & 21.9 & 47.9 & 28.2 \\

& SKCDF~\cite{zhang2025semantic} &
79.0 & 74.2 & 76.6 & 36.1 & 39.9 & 87.5 & 48.2 & 76.9 & 59.7 & 49.3 & 33.0 & 24.4 & 32.6 & 46.0 & 25.0 \\

& GenericSSL~\cite{wang2023towards} &
75.8 & 74.5 & 77.2 & 30.8 & 47.4 & 83.3 & 53.7 & 76.8 & 66.1 & 49.8 & 37.6 & 34.0 & 35.9 & 27.6 & 19.0\\

& GA-MagicNet~\cite{qi2024gradient} &
80.3 & 82.0 & 80.0 & 46.1 & 43.1 & \textbf{88.6} & 50.9 & 79.8 & 64.4 & 52.4 & 42.9 & 34.2 & 31.0 & \textbf{68.2} & 43.1 \\

& GA-CPS~\cite{qi2024gradient} &
71.3 & 62.0 & 59.0 & 35.2 & 38.2 & 76.1 & 50.7 & 62.2 & 58.8 & 43.9 & 44.0 & 28.8 & 29.0 & 62.6 & 41.6 \\
\midrule


& \textbf{DHC+SCDL} &
75.2 & 68.7 & 66.4 & 42.8 & 34.3 & 84.0 & 49.4 & 72.3 & 58.8 & 47.0 & 31.3 & 17.9 & 32.9 & 49.0 & 26.4 \\

& \textbf{GA-MagicNet+SCDL} &
79.7 & \textbf{84.0} & \textbf{84.4} &
\textbf{50.0} & 50.6 &
86.8 & 55.0 &
\textbf{84.6} & \textbf{70.9} &
55.6 & \textbf{46.3} &
\textbf{40.0} & 33.1 &
65.6 & 46.3 \\

& \textbf{GA-CPS+SCDL} &
\textbf{81.7} & 80.7 & 82.3 &
46.4 & \textbf{53.1} &
87.0 & \textbf{63.0} &
80.7 & 65.4 &
\textbf{56.8} & 39.1 &
35.2 & \textbf{39.1} &
66.8 & \textbf{46.4} \\

\bottomrule
\end{tabular*}

\endgroup
\end{table*}

\noindent\textbf{Semantic Anchor Alignment.}
Each class proxy $\mu_c$ is aligned with its corresponding semantic anchor using a cosine similarity-based loss:

\begin{equation}
\mathcal{L}_{\text{SAC}} = \frac{1}{C} \sum_{c=1}^{C} \big[ 1 - \cos(\mu_c, \text{anchor}_c) \big].
\end{equation}

This constraint guides proxies to encode class-specific semantic information while maintaining consistency across classes, including those with few samples.

\subsection{Training Objective}
\label{sec:objective}

The overall training objective combines the original semi-supervised segmentation loss with the proposed SCDL losses. Specifically, the baseline framework provides the standard segmentation objective, including supervised loss on labeled data and consistency or pseudo-label supervision on unlabeled data. SCDL is introduced as an auxiliary distribution-learning module without replacing the original training objective.

The total loss is formulated as:
\begin{equation}
\mathcal{L}_{total}
=
\mathcal{L}_{base}
+
\lambda_{E2P}\mathcal{L}_{E2P}
+
\lambda_{P2E}\mathcal{L}_{P2E}
+
\lambda_{SAC}\mathcal{L}_{SAC}.
\end{equation}

Here, $\mathcal{L}_{base}$ denotes the original loss of the selected semi-supervised baseline. $\mathcal{L}_{E2P}$ and $\mathcal{L}_{P2E}$ are the bidirectional alignment losses in CDBA, and $\mathcal{L}_{SAC}$ denotes the semantic anchor constraint. The coefficients $\lambda_{E2P}$, $\lambda_{P2E}$, and $\lambda_{SAC}$ control the contribution of each SCDL component. This formulation preserves the original segmentation learning process while using SCDL to regularize the class-conditional feature space and reduce class-frequency bias.

\begin{figure*}[!t]
\centering

\includegraphics[width=0.90\textwidth]{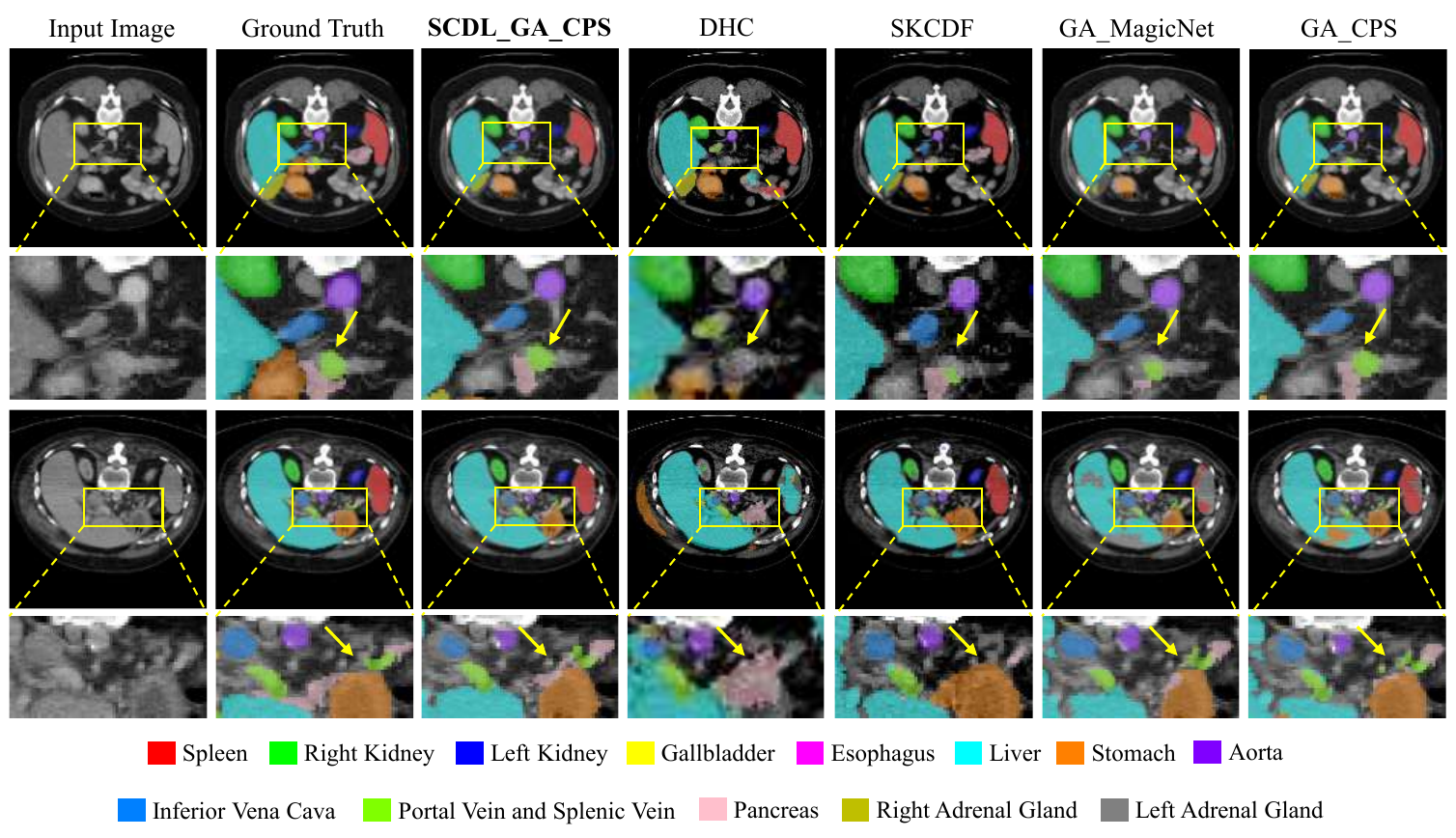}
\caption{\textbf{Qualitative comparison on the Synapse dataset.} 
Representative segmentation results are shown on two abdominal CT slices with zoomed-in views of challenging regions. Compared with DHC, SKCDF, GA-MagicNet, and GA-CPS, SCDL-GA-CPS produces masks that better match the ground-truth annotations, especially for small and anatomically adjacent organs such as the portal and splenic veins, pancreas, and inferior vena cava. These results suggest that SCDL helps reduce inter-class confusion under imbalanced organ distributions.
}
\label{fig:qualitative_synapse}


\includegraphics[width=0.90\textwidth]{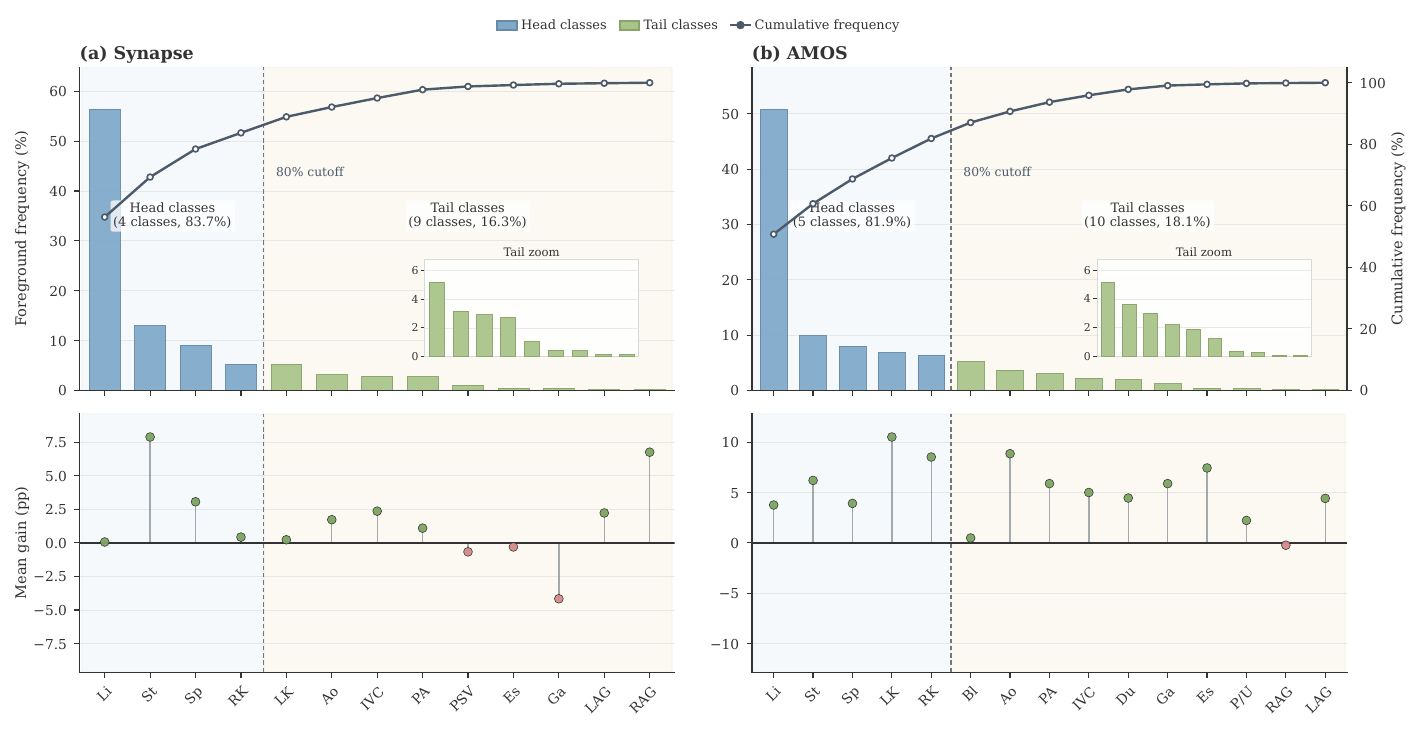}
\caption{\textbf{Head-tail class-frequency and Dice-gain analysis.} The upper panels show the foreground voxel frequency of each organ class in descending order, with the 80\% cumulative-frequency cutoff defining head and tail classes. On Synapse, four head classes account for 83.7\% of foreground voxels and nine tail classes for only 16.3\%; on AMOS, five head classes account for 81.9\% and ten tail classes for 18.1\%. The lower panels report the mean per-class Dice gain after adding SCDL, showing clear improvements for many low-frequency tail classes.
}
\label{fig:head_tail_frequency_improvement}

\end{figure*}

\begin{figure*}[!t]
\centering
\includegraphics[width=\textwidth]{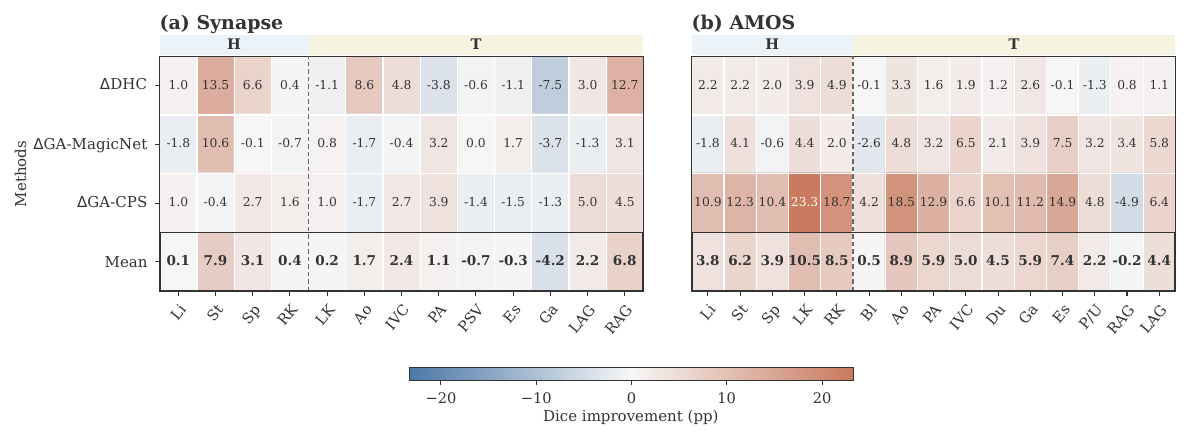}
\caption{\textbf{Per-class Dice improvement heatmap on Synapse and AMOS.}
Each cell represents the Dice improvement after adding SCDL to the corresponding baseline. 
$\Delta$DHC, $\Delta$GA-MagicNet, and $\Delta$GA-CPS denote the gains over DHC, GA-MagicNet, and GA-CPS, respectively. 
Classes are ordered by foreground voxel frequency, and the head-tail split follows Fig.~\ref{fig:head_tail_frequency_improvement}. 
The results show that SCDL brings larger gains to many tail classes, especially when coupled with stronger imbalance-aware baselines.  }
\label{fig:per_class_improvement_heatmap}
\end{figure*}

\section{Experiments}

\subsection{Dataset and Implementation Details}
We evaluate our method on two multi-organ CT datasets, Synapse\cite{landman2015btcv} and AMOS\cite{ji2022amos}. The Synapse dataset consists of 30 CT scans with 13 organ categories. The foreground classes include the spleen (Sp), right and left kidneys (RK, LK), gallbladder (Ga), esophagus (Es), liver (Li), stomach (St), aorta (Ao), inferior vena cava (IVC), portal and splenic veins (PSV), pancreas (PA), and right and left adrenal glands (RAG, LAG). Following the split used in DHC\cite{wang2023dhc} and GA\cite{qi2024gradient}, 20 scans are used for training, 4 for validation, and 6 for testing. The AMOS dataset comprises 360 subjects and 15 organ categories. Building upon Synapse, AMOS introduces three additional classes, prostate/uterus (P/U), duodenum (Du), and bladder (Bl), while excluding the portal and splenic veins (PSV). The dataset is split into 216 scans for training, 24 for validation, and 120 for testing. All experiments are conducted on an NVIDIA A40 GPU. The batch size is set to 4 for all experiments. The weight decay for the SCDL module is set to 1e-4. All other training configurations follow the corresponding baseline settings \cite{wang2023towards,wang2023dhc,qi2024gradient}. We report the Dice Similarity Coefficient (DSC,\%) and Average Surface Distance (ASD) for evaluation, where higher DSC and lower ASD indicate better performance.

\subsection{Comparisons with State-of-the-art Methods}

Table~\ref{tab:dsc_asd_summary} compares SCDL with representative semi-supervised segmentation methods under limited annotations. As a plug-in module, SCDL improves the mean DSC of all three baseline frameworks on both Synapse and AMOS, demonstrating its general applicability across different semi-supervised learning paradigms. On Synapse, DHC+SCDL increases the mean DSC from 47.88\% to 50.67\% and reduces ASD from 8.73 to 4.58, indicating that SCDL improves both region overlap and boundary accuracy for a relatively weak baseline. When coupled with stronger GA-based baselines, SCDL further improves the performance of GA-MagicNet and GA-CPS. In particular, GA-CPS+SCDL achieves the best semi-supervised result on Synapse, with 67.50\% mean DSC and 3.32 ASD, approaching the fully supervised VNet upper bound while using only 20\% labeled data.

On AMOS, where only 5\% annotations are available and the class-imbalance problem is more severe, SCDL also brings consistent DSC improvements. The gains are especially clear when SCDL is coupled with GA-based imbalance-aware baselines: GA-MagicNet+SCDL improves the mean DSC from 59.15\% to 62.16\% and reduces ASD from 8.66 to 5.65, while GA-CPS+SCDL increases the mean DSC from 50.90\% to 61.57\% and decreases ASD from 13.77 to 10.08. Although the ASD improvement is not uniform across all baseline combinations, SCDL reduces boundary errors in most settings while consistently improving segmentation accuracy. These results suggest that modeling semantic class distributions can effectively complement existing imbalance-aware semi-supervised frameworks.

Tables~\ref{tab:synapse_20p} and~\ref{tab:amos_5p} further report the per-class Dice scores on Synapse and AMOS, respectively. The improvements are not uniformly distributed across all organs, but are more evident on low-frequency or anatomically challenging categories. On Synapse, DHC+SCDL yields large gains on the stomach and right adrenal gland, improving Dice by 13.5 and 12.7 percentage points, respectively. GA-CPS+SCDL also improves several challenging structures, including the pancreas and adrenal glands. On AMOS, GA-CPS+SCDL brings substantial gains on multiple difficult organs, including the esophagus, pancreas, aorta, and kidneys, which are often affected by limited supervision, shape variability, or confusion with surrounding tissues.

To explicitly examine the relationship between these improvements and class imbalance, Fig.~\ref{fig:head_tail_frequency_improvement} visualizes the foreground voxel distribution, defines head and tail classes using the 80\% cumulative-frequency cutoff, and reports the average Dice gains after adding SCDL. The results show that many low-frequency classes benefit from SCDL. Fig.~\ref{fig:per_class_improvement_heatmap} further provides a per-baseline and per-class breakdown, where the improvements are especially visible on small or anatomically adjacent organs such as the pancreas, esophagus, and adrenal glands. Together with the qualitative comparison in Fig.~\ref{fig:qualitative_synapse}, these results indicate that SCDL helps alleviate class-distribution bias and improves class-level discrimination under limited annotations.

\subsection{Ablation Study}
\definecolor{HeaderGreen}{HTML}{87C1AA}
\begin{table}[!thbp]
\centering
\caption{Ablation study on the Synapse dataset.}
\label{tab:ablation_compare}

\begingroup
\fontsize{8}{9.2}\selectfont
\setlength{\tabcolsep}{2.2pt}
\renewcommand{\arraystretch}{1.05}

\begin{tabular*}{\columnwidth}{@{\extracolsep{\fill}}@{}
>{\centering\arraybackslash}m{0.42cm}
>{\centering\arraybackslash}m{1.35cm}
>{\centering\arraybackslash}m{0.85cm}
>{\centering\arraybackslash}m{0.75cm}
>{\centering\arraybackslash}m{0.85cm}
>{\centering\arraybackslash}m{0.85cm}
@{}}
\toprule
\rowcolor{HeaderGreen}
\# & Baseline & CDBA & SAC & DSC $\uparrow$ & ASD $\downarrow$ \\
\midrule
1 & \checkmark &  &  & 66.29 & 5.44 \\
2 & \checkmark & \checkmark &  & 66.77 & 6.24 \\
3 & \checkmark & \checkmark & \checkmark & \textbf{67.50} & \textbf{3.32} \\
\bottomrule
\end{tabular*}

\endgroup
\end{table}
To validate the effectiveness of the two core components in SCDL, we conduct an ablation study on Synapse using 20\% labeled data, as shown in Table~\ref{tab:ablation_compare}. The baseline achieves 66.29\% mean Dice and 5.44 ASD. Adding CDBA introduces class-conditional distributional constraints that improve region-level prediction consistency, leading to a +0.48\% gain in mean Dice. However, ASD increases by +0.80, indicating that distribution-level alignment alone does not reliably improve boundary geometry. After further incorporating SAC, the full model attains the best performance with 67.50\% mean Dice (+0.73\% over CDBA) and a markedly reduced ASD of 3.32 ($\downarrow 2.92$). These results suggest that semantic constraints from labeled regions enhance the stability and semantic consistency of proxies, thereby improving geometric boundary quality more reliably.

\section{Conclusion}
Semi-supervised methods often struggle with pixel-level long-tailed imbalance, where dominant large structures and pseudo-labeling with consistency regularization bias learning, destabilizing small-structure segmentation and blurring class boundaries. We propose SCDL, a class distribution debiasing framework for semi-supervised medical segmentation. It first uses bidirectional alignment to stabilize minority-class proxies, then semantic anchoring refines them toward labeled centers, preventing drift to majority classes. Across multiple datasets, SCDL consistently yields more balanced solutions with improved boundary precision on anatomically scarce regions.

\bibliographystyle{IEEEtran}
\bibliography{references}

\end{document}